\documentclass[11pt,a4paper]{article}
\usepackage[hyperref]{acl2019}
\usepackage{times}
\usepackage{latexsym}

\usepackage{booktabs}
\usepackage{multirow}
\usepackage{graphicx}
\usepackage{caption}
\usepackage[normalem]{ulem}
\usepackage[utf8]{inputenc}
\usepackage{dingbat}
\usepackage{wrapfig}
\usepackage{amssymb}
\usepackage{amsfonts}
\usepackage{amsmath}
\usepackage{amssymb}
\usepackage{amsthm}
\usepackage{mathrsfs}
\usepackage{adjustbox}
\usepackage{bbold}
\usepackage[font=small]{caption}
\usepackage[linesnumbered,ruled]{algorithm2e}
\usepackage{url}

\DeclareMathOperator*{\argmax}{argmax}

\usepackage{enumitem}
\usepackage{wrapfig,lipsum,booktabs}

\newcommand{\dis}{Target Conditioned Sampling}
\newcommand{\disab}{TCS}

\aclfinalcopy 
\def\aclpaperid{1290} 

\title{\dis: \\ Optimizing Data Selection for Multilingual Neural Machine Translation}

\author{Xinyi Wang\\
  Language Technologies Institute \\
  Carnegie Mellon University \\
  \texttt{xinyiw1@cs.cmu.edu} \\\And
  Graham Neubig \\
  Language Technologies Institute \\
  Carnegie Mellon University \\
  \texttt{gneubig@cs.cmu.edu} \\}

\date{}

\begin{document}
\maketitle
\begin{abstract}
To improve low-resource Neural Machine Translation~(NMT) with multilingual corpora, training on the most related high-resource language only is often more effective than using all data available~\cite{rapid_adapt_nmt}. However, it is possible that an intelligent data selection strategy can further improve low-resource NMT with data from other auxiliary languages. In this paper, we seek to construct a sampling distribution over all multilingual data, so that it minimizes the training loss of the low-resource language.
Based on this formulation, we propose an efficient algorithm, \dis~(\disab), which first samples a target sentence, and then conditionally samples its source sentence. Experiments show that \disab~brings significant gains of up to 2 BLEU on three of four languages we test, with minimal training overhead.    
\end{abstract}

\section{\label{sec:intro}Introduction}




Multilingual NMT has led to impressive gains in translation accuracy of low-resource languages~(LRL)~\citep{multi_nmt_adapt,multi_nmt_shared_attn,universal_nmt,rapid_adapt_nmt,multi_nmt_bpe_share}.
Many real world datasets provide sentences that are multi-parallel, with the same content in a variety of languages.
Examples include TED~\cite{ted_pretrain_emb}, Europarl~\cite{europarl}, and many others~\cite{opus}. 
These datasets open up the tantalizing prospect of training a system on many different languages to improve accuracy, but previous work has found methods that use only a single related~(HRL) often out-perform systems trained on all available data~\citep{rapid_adapt_nmt}.
In addition, because the resulting training corpus is smaller, using a single language is also substantially faster to train, speeding experimental cycles~\citep{rapid_adapt_nmt}.
In this paper, we go a step further and ask the question: can we design an intelligent data selection strategy that allows us to choose the most relevant multilingual data to further boost NMT performance and training speed for LRLs?  

Prior work has examined data selection from the view of domain adaptation, selecting good training data from out-of-domain text to improve in-domain performance.
In general, these methods select data that score above a preset threshold according to some metric, such as the difference between in-domain and out-of-domain language models~\cite{domain_adapt_data_select,lm_data_select} or sentence embedding similarity~\cite{wang-EtAl:2017:Short3}.
Other works use all the data but weight training instances by domain similarity~\cite{chen-EtAl:2017:NMT}, or sample subsets of training data at each epoch~\cite{dynamic_data_select}.
However, none of these methods are trivially applicable to multilingual parallel datasets, which usually contain many \textit{different} languages from the \textit{same} domain.
Moreover, most of these methods need to pretrain language models or NMT models with a reasonable amount of data, and accuracy can suffer in low-resource settings like those encountered for LRLs \cite{duh-EtAl:2013:Short}. 

In this paper, we create a mathematical framework for data selection in multilingual MT that selects data from \textit{all} languages, such that minimizing the training objective over the sampled data approximately minimizes the loss of the LRL MT model.
The formulation leads to an simple, efficient, and effective algorithm that first samples a target sentence and then conditionally samples which of several source sentences to use for training.
We name the method \dis~(\disab).
We also propose and experiment with several design choices for \disab, which are especially effective for LRLs.
On the TED multilingual corpus \cite{ted_pretrain_emb}, \disab~leads to large improvements of up to 2 BLEU on three of the four languages we test, and no degradation on the fourth, with only slightly increased training time. To our knowledge, this is the first successful application of data selection to multilingual NMT.

\section{\label{sec:method}Method}
\subsection{Multilingual Training Objective}

First, in this section we introduce our problem formally, where we use the upper case letters $X$, $Y$ to denote the random variables, and the corresponding lower case letters $x$, $y$ to denote their actual values. 
Suppose our objective is to learn parameters $\theta$ of a translation model from a source language $s$ into target language $t$. 
Let $x$ be a source sentence from $s$, and $y$ be the equivalent target sentence from $t$, 
given loss function $\mathcal{L}(x, y;\theta)$
our objective is to find optimal parameters $\theta^*$ that minimize:
\begin{equation}
  \label{eqn:objective}
  \begin{aligned}
    \mathbb{E}_{x, y \sim P_S(X, Y)} [\mathcal{L}(x, y;\theta)]
  \end{aligned}
\end{equation}
where $P_s(X, Y)$ is the data distribution of $s$-$t$ parallel sentences.

Unfortunately, we do not have enough data to accurately estimate $\theta^*$, but instead we have a multilingual corpus of parallel data from languages $\{s_1, S_2, ..., S_n\}$ all into $t$.
Therefore, we resort to multilingual training to facilitate the learning of $\theta$. Formally, we want to construct a distribution $Q(X, Y)$ with support over $s_1, s_2,...,s_n$-$T$ to augment the $s$-$t$ data with samples from $Q$ during training. Intuitively, a good $Q(X,Y)$ will have an expected loss
\begin{equation}
  \label{eqn:objective_q}
  \begin{aligned}
    \mathbb{E}_{x, y \sim Q(X, Y)} [\mathcal{L}(x, y;\theta)]
  \end{aligned}
\end{equation}
that is correlated with Eqn \ref{eqn:objective} over the space of all $\theta$, so that training over data sampled from $Q(X,Y)$ can facilitate the learning of $\theta$. Next, we explain a version of $Q(X, Y)$ designed to promote efficient multilingual training.

\subsection{\dis}
We argue that the optimal $Q(X,Y)$ should satisfy the following two properties. 

First, $Q(X, Y)$ and $P_s(X, Y)$ should be \textbf{target invariant};
the marginalized distributions $Q(Y)$ and $P_s(Y)$ should match as closely as possible:
    \begin{equation}
        \begin{aligned}
        \label{eqn:Q2}
        Q(Y) \approx P_s(Y)
        \end{aligned}
    \end{equation}
This property ensures that Eqn \ref{eqn:objective} and Eqn \ref{eqn:objective_q} are optimizing towards the same target $Y$ distribution.



Second, to have Eqn \ref{eqn:objective_q} correlated with Eqn \ref{eqn:objective} over the space of all $\theta$, we need $Q(X,Y)$ to be correlated with $P_s(X, Y)$, which can be loosely written as  
    \begin{equation}
        \begin{aligned}
        \label{eqn:Q1}
        Q(X, Y) \approx P_{s} (X, Y).
        \end{aligned}
    \end{equation}
    Because we also make the target invariance assumption in Eqn \ref{eqn:Q2},
    \begin{align}
        \frac{Q(X, Y)}{Q(Y)} & \approx \frac{P_{s}(X, Y)}{P_{s}(Y)} \\
        Q(X|Y) & \approx P_{s}(X|Y).
        \label{eqn:Q3}
    \end{align}
    We call this approximation of $P_{s}(X|Y)$ by $Q(X|Y)$ \textbf{conditional source invariance}.
 Based on these two assumptions, we define \dis~(\disab), a training framework that first samples $y \sim Q(Y)$, and then conditionally samples $x \sim Q(X|y)$ during training.
 Note $P_{s}(X|Y=y)$ is the optimal back-translation distribution, which implies that back-translation \cite{sennrich} is a particular instance of \disab.

 Of course, we do not have enough $s$-$t$ parallel data to obtain a good estimate of the \emph{true} back-translation distribution $P_{s}(X|y)$ (otherwise, we can simply use that data to learn $\theta$).
 However, we posit that even a small amount of data is sufficient to construct an adequate data selection policy $Q(X|y)$ to sample the sentences $x$ from multilingual data for training.
 Thus, the training objective that we optimize is
\begin{equation}
  \label{eqn:importance_4}
  \begin{aligned}
    \mathbb{E}_{y \sim Q(Y)} \mathbb{E}_{x \sim Q(X|y)} \left[ \mathcal{L}(x, y; \theta) \right]
  \end{aligned}
\end{equation}
Next, in Section~\ref{sec:choosing_distributions}, we discuss the choices of $Q(Y)$ and $Q(X|y)$.

\subsection{\label{sec:choosing_distributions}Choosing the Sampling Distributions}

\paragraph{Choosing $Q(Y)$.} Target invariance requires that we need $Q(Y)$ to match $P_s(Y)$, which is the distribution over the target of $s$-$t$. We have parallel data from multiple languages $s_1, s_2, ..., s_n$, all into $t$. Assuming no systematic inter-language distribution differences, a uniform sample of a target sentence $y$ from the multilingual data can approximate $P_s(Y)$. We thus only need to sample $y$ uniformly from the union of all extra data. 

\paragraph{Choosing $Q(X|y)$.}
Choosing $Q(X|y)$ to approximate $P_s(X|y)$ is more difficult, and there are a number of methods could be used to do so.
To do so, we note that conditioning on the same target $y$ and restricting the support of $P_{s}(X|y)$ to the sentences that translate into $y$ in at least one of $s_i$-$t$, $P_{s}(X=x|y)$ simply measures how likely $x$ is in $s$.
We thus define a heuristic function $\text{sim}(x,s)$ that approximates the probability that $x$ is a sentence in $s$, and follow the data augmentation objective in~\citet{switchout} in defining this probability according to
\begin{align}
  Q^*(x|y) = \frac{\exp{(\text{sim}(x, s)/\tau)}}{\sum_{x'} \exp{(\text{sim}(x', s)/\tau)}}
\end{align}
where is a temperature parameter that adjusts the peakiness of the distribution.


\subsection{\label{subsec:alg}Algorithms}
 The formulation of $Q(X, Y)$ allows one to sample multilingual data with the following algorithm:
\setlist{nolistsep}
\begin{enumerate}[noitemsep]
    \item Select the target $y$ based on $Q(y)$. In our case we can simply use the uniform distribution. 
    \item Given the target $y$, gather all data $(x_i, y) \in$ $s_1,s_2,...s_n$-$t$ and calculate $\text{sim}(x_i, s)$
   \item Sample $(x_i, y)$ based on $Q(X|y)$ 
\end{enumerate}
The algorithm requires calculating $Q(X|y)$ repeatedly during training. To reduce this overhead, we propose two strategies for implementation: 1) \textbf{Stochastic}: compute $Q(X|y)$ before training starts, and dynamically sample each minibatch using the precomputed $Q(X|y)$; 2) \textbf{Deterministic}: compute $Q(X|y)$ before training starts and select $x' = \argmax_{x} Q(x|y)$ for training.
The deterministic method is equivalent to setting $\tau$, the degree of diversity in $Q(X|y)$, to be 0.

\subsection{\label{subsec:sim}Similarity Measure}
In this section, we define two formulations of the similarity measure $\text{sim}(s, x)$, which is essential for constructing $Q(X|y)$. Each of the similarity measures can be calculated at two granularities: 1) language level, which means we calculate one similarity score for each language based on all of its training data; 2) sentence level, which means we calculate a similarity score for each sentence in the training data.
\paragraph{Vocab Overlap} provides a crude measure of surface form similarity between two languages. It is efficient to calculate, and is often quite effective, especially for low-resource languages. Here we use the number of character $n$-grams that two languages share to measure the similarity between the two languages. 

We can calculate the language-level similarity between $S_i$ and $S$ 
\begin{align*}
    \text{sim}_{\text{vocab-lang}}(s_i, s) = \frac{| \text{vocab}_k(s) \cap \text{vocab}_k(s_i)|}{k}
\end{align*}
$\text{vocab}_k(\cdot)$ represents the top $k$ most frequent character $n$-grams in the training data of a language. Then we can assign the same language-level similarity to all the sentences in $s_i$.

This can be easily extended to the sentence level by replacing $\text{vocab}_k(s_i)$ to the set of character $n$-grams of all the words in the sentence $x$.
\setlength{\abovedisplayskip}{1pt}
\setlength{\belowdisplayskip}{1pt}

\paragraph{Language Model} trained on $s$ can be used to calculate the probability that a data sequence belongs to $s$. Although it might not perform well if $s$ does not have enough training data, it may still be sufficient for use in the \disab~algorithm. The language-level metric is defined as 
\begin{align*}
    \text{sim}_{\text{LM-lang}}(s_i, s) = \text{exp}\left( \frac{\sum_{c_i \in s_i}\text{NLL}_{s} (c_i)}{|c_i \in s_i|}\right)
\end{align*}
where $\text{NLL}_s(\cdot)$ is negative log likelihood of a character-level LM trained on data from $s$.
Similarly, the corresponding sentence level metric is the LM probability over each sentence $x$. 


\section{\label{sec:exp}Experiment}
\subsection{Dataset and Baselines}
\begin{table}[]
  \centering
   \resizebox{0.45\textwidth}{!}{
  \begin{tabular}{c|ccc|cc}
  \toprule
  \textbf{LRL} & \textbf{Train} & \textbf{Dev} & \textbf{Test} & \textbf{HRL} & \textbf{Train} \\
  \midrule
  aze & 5.94k & 671 & 903 & tur & 182k \\
  bel & 4.51k & 248 & 664 & rus & 208k \\
  glg & 10.0k & 682 & 1007 & por & 185k \\
  slk & 61.5k & 2271 & 2445 & ces & 103k\\
  \bottomrule
  \end{tabular}}
  \caption{\label{tab:data}Statistics of our datasets.}
\end{table} 

We use the 58-language-to-English TED dataset~\citep{ted_pretrain_emb}. 
Following the setup in prior work~\citep{ted_pretrain_emb,rapid_adapt_nmt}, we use three low-resource languages Azerbaijani (aze), Belarusian (bel), Galician (glg) to English,
and a slightly higher-resource dataset, Slovak (slk) to English.

We use multiple settings for baselines: 1) Bi: each LRL is paired with its related HRL, following~\citet{rapid_adapt_nmt}. The statistics of the LRL and their corresponding HRL are listed in Table \ref{tab:data}; 2) All: we train a model on all 58 languages; 3) Copied: following~\citet{copied_mono}, we use the union of all English sentences as monolingual data by copying them to the source side.

\subsection{Experiment Settings}
 A standard sequence-to-sequence~\cite{seq2seq} NMT model with attention is used for all experiments. Byte Pair Encoding~(BPE)~\cite{bpe,sentencepiece} with vocabulary size of 8000 is applied for each language individually. Details of other hyperparameters can be found in  Appendix~\ref{subsec:hyperparam}.

\subsection{\label{subsec:result} Results}
\begin{table}[t]
    \centering
    \resizebox{0.5\textwidth}{!}{
    \begin{tabular}{cc|cccc}
    \toprule
    \textbf{Sim} & \textbf{Method} & \textbf{aze} & \textbf{bel} & \textbf{glg} & \textbf{slk} \\
    \midrule
    - & Bi & 10.35 & 15.82 & 27.63 & 26.38 \\
    - & All & 10.21 & 17.46 & 26.01 & 26.64 \\
    - & copied & 9.54 & 13.88 & 26.24 & 26.77 \\
    \midrule
    Back-Translate & \disab & 7.79 & 11.50 & 27.45 & 28.44 \\
    \midrule
    LM-sent & \disab-D & 10.34 & 14.68 & 27.90 & 27.29 \\
    LM-sent & \disab-S & 10.95 & 17.15 & 27.91 & 27.24 \\
    \midrule
    LM-lang & \disab-D & 10.76 & 14.97 & 27.92 & 28.40 \\
    LM-lang & \disab-S & $\mathbf{11.47^*}$ & 17.61 & 28.53 & $\mathbf{28.56^*}$ \\
    \midrule
    Vocab-sent & \disab-D & 10.68 & 16.13 & 27.29 & 27.03 \\
    Vocab-sent & \disab-S & 11.09 & 16.30 & 28.36 & 27.01 \\
    \midrule
    Vocab-lang & \disab-D & 10.58 & 16.32 & 28.17 & 28.27 \\
    Vocab-lang & \disab-S & $\mathbf{11.46^*}$ & $\mathbf{17.79}$ & $\mathbf{29.57^*}$ & 28.45 \\
    \bottomrule
    \end{tabular}}
    \caption{BLEU scores on four languages. Statistical significance~\cite{significance_test} is indicated with $*$~($p < 0.001$), compared with the best baseline.} 
    \label{tab:results}
\end{table}
We test both the Deterministic~(\disab-D) and Stochastic~(\disab-S) algorithms described in Section~\ref{subsec:alg}. For each algorithm, we experiment with the similarity measures introduced in Section~\ref{subsec:sim}. The results are listed in Table~\ref{tab:results}.

Of all the baselines, Bi in general has the best performance, while All, which uses all the data and takes much longer to train, generally hurts the performance. This is consistent with findings in prior work~\cite{rapid_adapt_nmt}. Copied is only competitive for slk, which indicates the gain of \disab~is not simply due to extra English data. 
 
 \disab-S combined with the language-level similarity achieves the best performance for all four languages, improving around 1 BLEU over the best baseline for aze, and around 2 BLEU for glg and slk. For bel, \disab~leads to no degradation while taking much less training time than the best baseline All.


\paragraph{\disab-D vs. \disab-S.} Both algorithms, when using document-level similarity, improve over the baseline for all languages. \disab-D is quite effective without any extra sampling overhead. \disab-S outperforms \disab-D for all experiments, indicating the importance of diversity in the training data. 

\paragraph{Sent. vs. Lang.} For all experiments, language-level outperforms the sentence-level similarity. This is probably because language-level metric provides a less noisy estimation, making $Q(x|y)$ closer to $P_{s}(x|y)$.

\paragraph{LM vs. Vocab.} In general, the best performing methods using LM and Vocab are comparable, except for glg, where Vocab-lang outperforms LM-lang by 1 BLEU. Slk is the only language where LM outperformed Vocab in all settings, probably because it has the largest amount of data to obtain a good language model. These results show that easy-to-compute language similarity features are quite effective for data selection in low-resource languages.

\paragraph{Back-Translation}

\disab~constructs $Q(X|y)$ to sample augmented multilingual data, when the LRL data cannot estimate a good back-translation model. Here we confirm this intuition by replacing the $Q(X|y)$ in \disab~with the back-translations generated by the model trained on the LRLs. To make it comparable to Bi, we use the sentence from the LRL and its most related HRL if there is one for the sampled $y$, but use the back-translated sentence otherwise. 
Table \ref{tab:results} shows that for slk, back-translate achieves comparable results with the best similarity measure, mainly because slk has enough data to get a reasonable back-translation model. However, it performs much worse for aze and bel, which have the smallest amount of data.
 
 \subsection{\label{subsec:sde}Effect on SDE}
 
\begin{table}[t]
    \centering
     \resizebox{0.45\textwidth}{!}{
    \begin{tabular}{cc|cccc}
    \toprule
    \textbf{Sim} & \textbf{Model} & \textbf{aze} & \textbf{bel} & \textbf{glg} & \textbf{slk} \\
    \midrule
    - & Bi & 11.87 & 18.03 & 28.70 & 26.77 \\
    - & All & 10.87 & 17.77 & 25.49 & 26.28  \\
    - & copied & 10.74 & 17.19 & 29.75 & 27.81 \\
    \midrule
    LM-lang & \disab-D & 11.97 & 17.17 & 30.10 & 28.78 \\
    LM-lang & \disab-S & $\mathbf{12.55^\dagger}$ & 17.23 & 30.69 & 28.95 \\
    \midrule
    Vocab-lang & \disab-D & 12.30 & 18.96 & $\mathbf{31.10^*}$ & $\mathbf{29.35^*}$ \\
    Vocab-lang & \disab-S & 12.37 & $\mathbf{19.83^\dagger}$ & 30.94 & 29.00 \\
    \bottomrule
    \end{tabular}}
    \caption{BLEU scores using SDE as word encoding. Statistical significance is indicated with $*$~($p < 0.001$) and $\dagger$~($p < 0.05$), compared with the best baseline.} 
    \label{tab:sde}
\end{table}

To ensure that our results also generalize to other models, specifically ones that are tailored for better sharing of information across languages, we also test \disab~on a slightly different multilingual NMT model using soft decoupled encoding (SDE; \citet{sde}), a word encoding method that assists lexical transfer for multilingual training.
The results are shown in Table \ref{tab:sde}.
Overall the results are stronger, but the best \disab~model outperforms the baseline by 0.5 BLEU for aze, and around 2 BLEU for the rest of the three languages, suggesting the orthogonality of data selection and better multilingual training methods.

\subsection{Effect on Training Curves}
\begin{figure}[t]
\begin{center}
  \includegraphics[width=0.45\columnwidth]{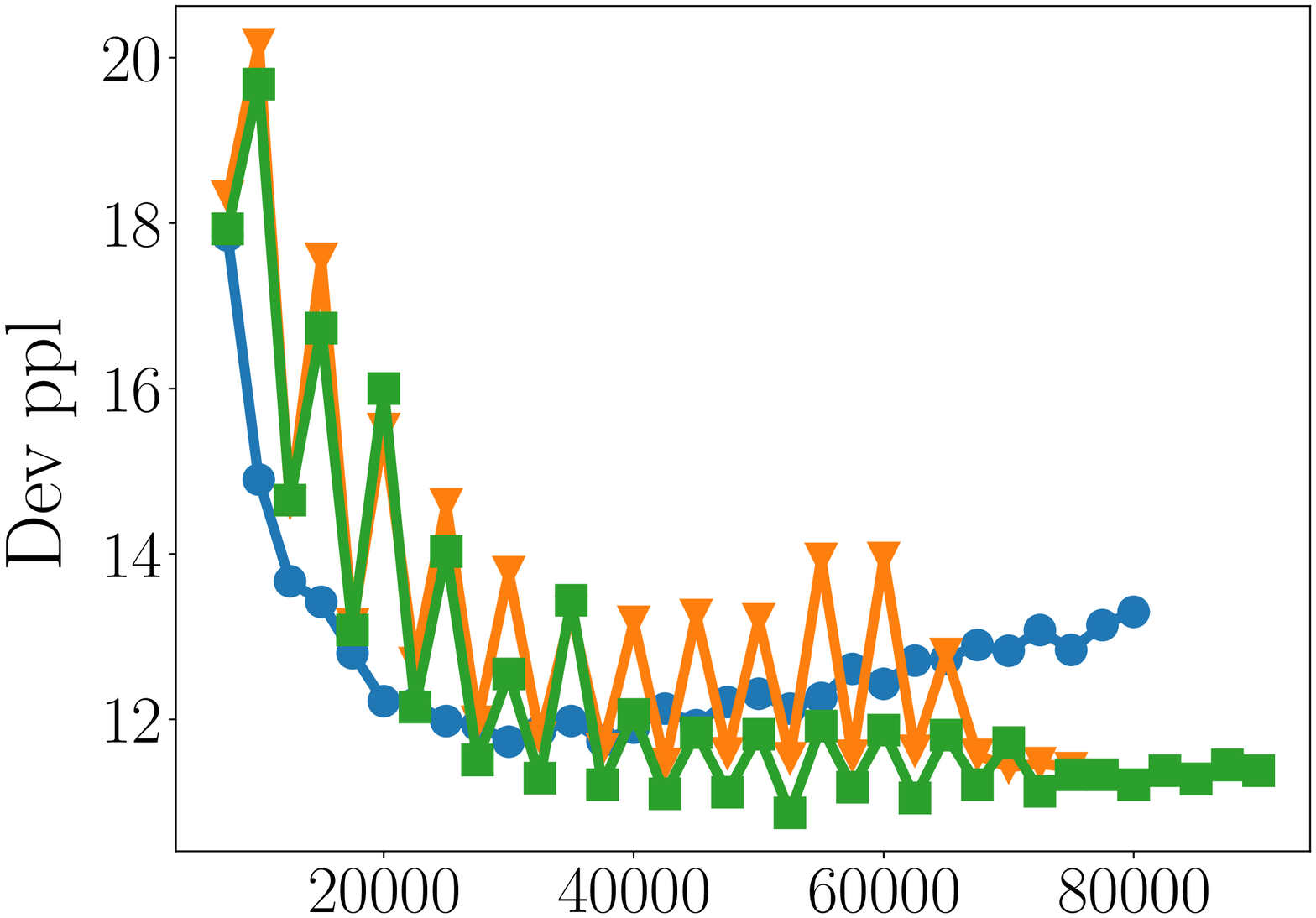}
  \includegraphics[width=0.42\columnwidth]{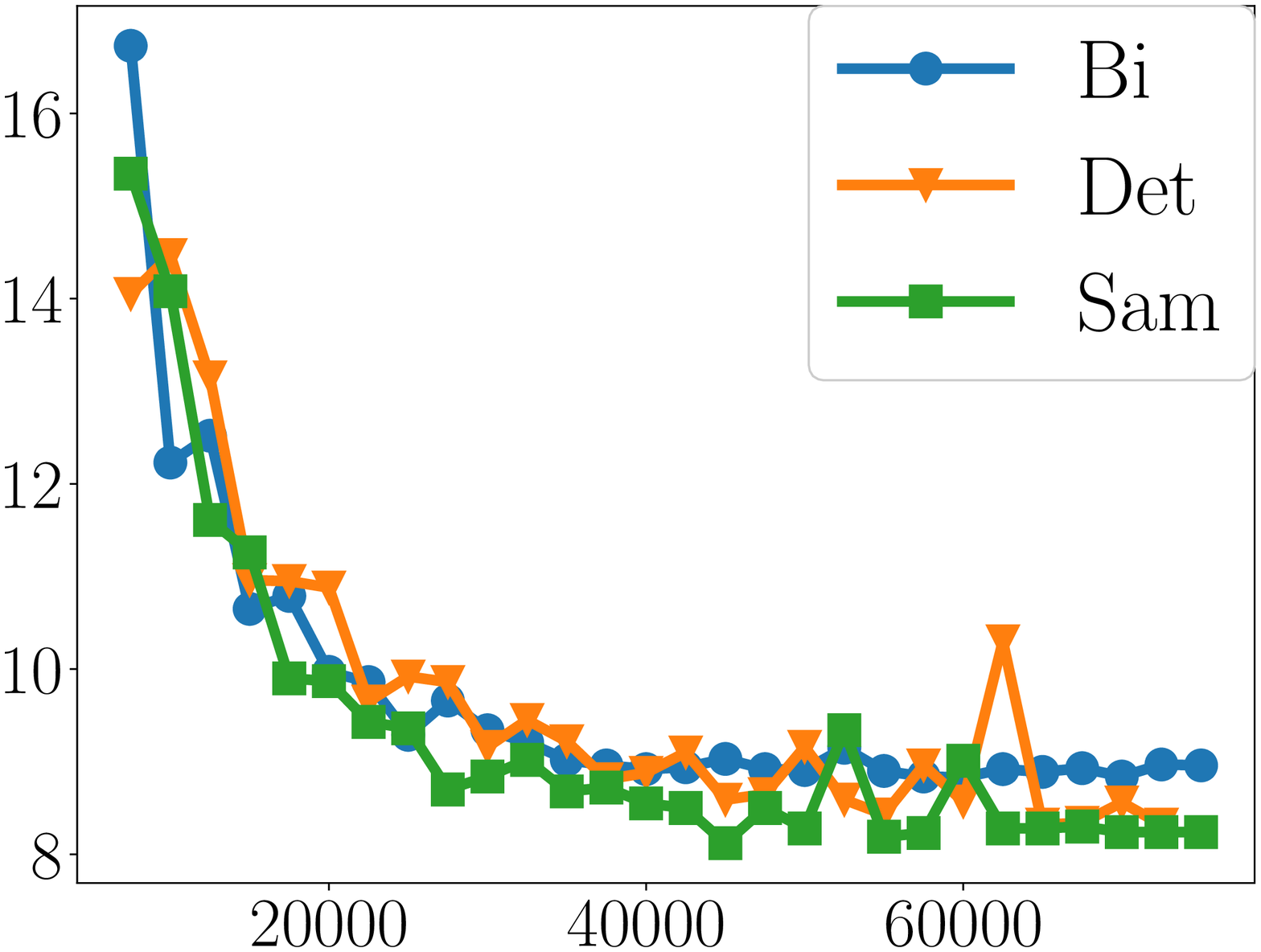}
  \includegraphics[width=0.45\columnwidth]{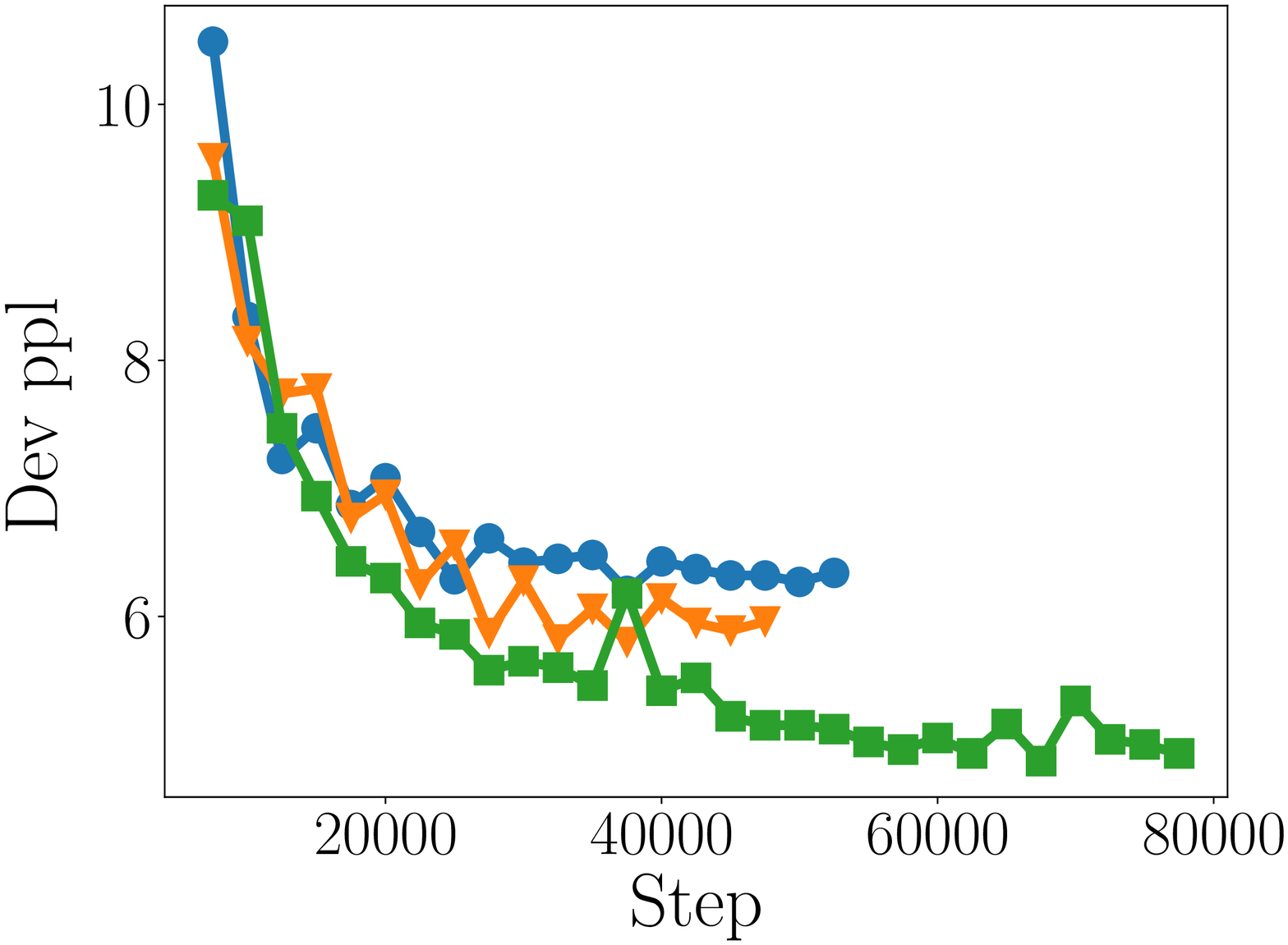}
  \includegraphics[width=0.42\columnwidth]{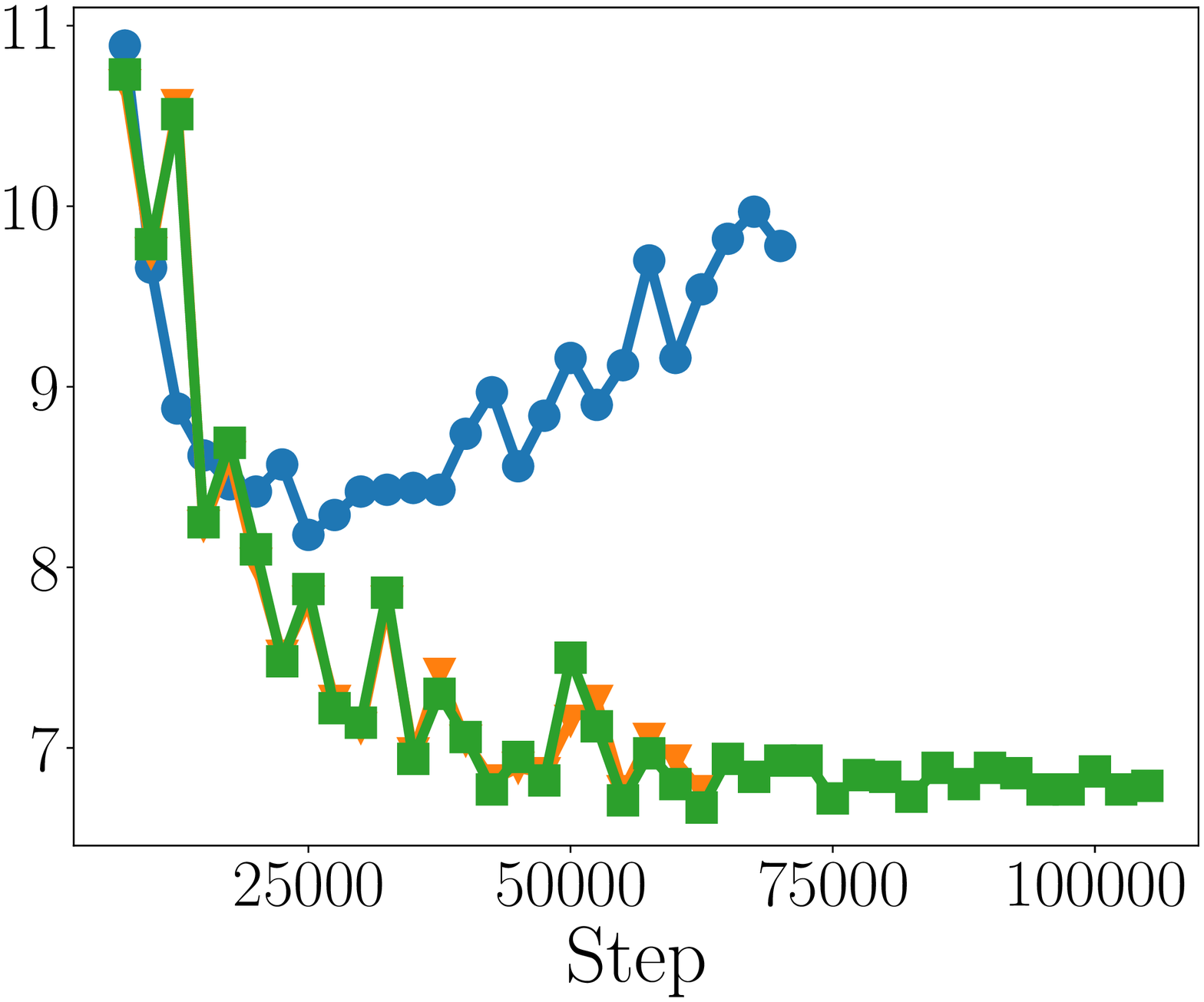}
  \captionof{figure}{\label{fig:converge}Development set perplexity vs. training steps. \textit{Top left}: aze. \textit{Top right}: bel. \textit{Bottom left}: glg. \textit{Bottom right}: slk.}
\end{center}
\end{figure}
In Figure \ref{fig:converge}, we plot the development perplexity of all four languages during training. Compared to Bi, \disab~always achieves lower development perplexity, with only slightly more training steps. Although using all languages, \disab~is able to decrease the development perplexity at similar rate as Bi. This indicates that \disab~is effective at sampling helpful multilingual data for training NMT models for LRLs.  
\section{\label{sec:con}Conclusion}
We propose \dis~(\disab), an efficient data selection framework for multilingual data by constructing a data sampling distribution that facilitates the NMT training of LRLs. \disab~brings up to 2 BLEU improvements over strong baselines with only slight increase in training time.

\section*{Acknowledgements}
The authors thank Hieu Pham and Zihang Dai for helpful discussions and comments on the paper. We also thank Paul Michel, Zi-Yi Dou, and Calvin McCarter for proofreading the paper.
This material is based upon work supported in part by the Defense Advanced Research Projects Agency Information Innovation Office (I2O) Low Resource Languages for Emergent Incidents (LORELEI) program under Contract No. HR0011-15-C0114. The views and conclusions contained in this document are those of the authors and should not be interpreted as representing the official policies, either expressed or implied, of the U.S. Government. The U.S. Government is authorized to reproduce and distribute reprints for Government purposes notwithstanding any copyright notation here on.

\bibliography{main}

\begin{thebibliography}{21}
\expandafter\ifx\csname natexlab\endcsname\relax\def\natexlab#1{#1}\fi

\bibitem[{Axelrod et~al.(2011)Axelrod, He, and Gao}]{domain_adapt_data_select}
Amittai Axelrod, Xiaodong He, and Jianfeng Gao. 2011.
\newblock Domain adaptation via pseudo in-domain data selection.
\newblock In \emph{EMNLP}.

\bibitem[{Chen et~al.(2017)Chen, Cherry, Foster, and
  Larkin}]{chen-EtAl:2017:NMT}
Boxing Chen, Colin Cherry, George Foster, and Samuel Larkin. 2017.
\newblock Cost weighting for neural machine translation domain adaptation.
\newblock In \emph{WMT}.

\bibitem[{Clark et~al.(2011)Clark, Dyer, Lavie, and Smith}]{significance_test}
Jonathan Clark, Chris Dyer, Alon Lavie, and Noah Smith. 2011.
\newblock Better hypothesis testing for statistical machine translation:
  Controlling for optimizer instability.
\newblock In \emph{ACL}.

\bibitem[{Currey et~al.(2017)Currey, Miceli~Barone, and Heafield}]{copied_mono}
Anna Currey, Antonio~Valerio Miceli~Barone, and Kenneth Heafield. 2017.
\newblock Copied monolingual data improves low-resource neural machine
  translation.
\newblock In \emph{WMT}.

\bibitem[{Duh et~al.(2013)Duh, Neubig, Sudoh, and
  Tsukada}]{duh-EtAl:2013:Short}
Kevin Duh, Graham Neubig, Katsuhito Sudoh, and Hajime Tsukada. 2013.
\newblock Adaptation data selection using neural language models: Experiments
  in machine translation.
\newblock In \emph{ACL}.

\bibitem[{Firat et~al.(2016)Firat, Cho, and Bengio}]{multi_nmt_shared_attn}
Orhan Firat, Kyunghyun Cho, and Yoshua Bengio. 2016.
\newblock Multi-way, multilingual neural machine translation with a shared
  attention mechanism.
\newblock \emph{NAACL}.

\bibitem[{Gu et~al.(2018)Gu, Hassan, Devlin, and Li}]{universal_nmt}
Jiatao Gu, Hany Hassan, Jacob Devlin, and Victor O.~K. Li. 2018.
\newblock Universal neural machine translation for extremely low resource
  languages.
\newblock \emph{NAACL}.

\bibitem[{Koehn(2005)}]{europarl}
Philipp Koehn. 2005.
\newblock Europarl: A parallel corpus for statistical machine translation.

\bibitem[{Kudo and Richardson(2018)}]{sentencepiece}
Taku Kudo and John Richardson. 2018.
\newblock Sentencepiece: A simple and language independent subword tokenizer
  and detokenizer for neural text processing.
\newblock In \emph{EMNLP}.

\bibitem[{Moore and Lewis(2010)}]{lm_data_select}
Robert~C. Moore and William~D. Lewis. 2010.
\newblock Intelligent selection of language model training data.
\newblock In \emph{ACL}.

\bibitem[{Neubig and Hu(2018)}]{rapid_adapt_nmt}
Graham Neubig and Junjie Hu. 2018.
\newblock Rapid adaptation of neural machine translation to new languages.
\newblock \emph{EMNLP}.

\bibitem[{Nguyen and Chiang(2018)}]{multi_nmt_bpe_share}
Toan~Q. Nguyen and David Chiang. 2018.
\newblock Transfer learning across low-resource, related languages for neural
  machine translation.
\newblock In \emph{NAACL}.

\bibitem[{Qi et~al.(2018)Qi, Sachan, Felix, Padmanabhan, and
  Neubig}]{ted_pretrain_emb}
Ye~Qi, Devendra~Singh Sachan, Matthieu Felix, Sarguna Padmanabhan, and Graham
  Neubig. 2018.
\newblock When and why are pre-trained word embeddings useful for neural
  machine translation?
\newblock \emph{NAACL}.

\bibitem[{Sennrich et~al.(2016)Sennrich, Haddow, and Birch}]{bpe}
Rico Sennrich, Barry Haddow, and Alexandra Birch. 2016.
\newblock Neural machine translation of rare words with subword units.
\newblock In \emph{ACL}.

\bibitem[{Sutskever et~al.(2014)Sutskever, Vinyals, and Le}]{seq2seq}
Ilya Sutskever, Oriol Vinyals, and Quoc~V. Le. 2014.
\newblock Sequence to sequence learning with neural networks.
\newblock In \emph{NIPS}.

\bibitem[{Tiedemann(2012)}]{opus}
Jörg Tiedemann. 2012.
\newblock Parallel data, tools and interfaces in opus.
\newblock In \emph{LREC}.

\bibitem[{Wang et~al.(2017)Wang, Finch, Utiyama, and
  Sumita}]{wang-EtAl:2017:Short3}
Rui Wang, Andrew Finch, Masao Utiyama, and Eiichiro Sumita. 2017.
\newblock Sentence embedding for neural machine translation domain adaptation.
\newblock In \emph{ACL}.

\bibitem[{Wang et~al.(2019)Wang, Pham, Arthur, and Neubig}]{sde}
Xinyi Wang, Hieu Pham, Philip Arthur, and Graham Neubig. 2019.
\newblock Multilingual neural machine translation with soft decoupled encoding.
\newblock In \emph{ICLR}.

\bibitem[{Wang et~al.(2018)Wang, Pham, Dai, and Neubig}]{switchout}
Xinyi Wang, Hieu Pham, Zihang Dai, and Graham Neubig. 2018.
\newblock Switchout: an efficient data augmentation algorithm for neural
  machine translation.
\newblock In \emph{EMNLP}.

\bibitem[{van~der Wees et~al.(2017)van~der Wees, Bisazza, and
  Monz}]{dynamic_data_select}
Marlies van~der Wees, Arianna Bisazza, and Christof Monz. 2017.
\newblock Dynamic data selection for neural machine translation.
\newblock In \emph{EMNLP}.

\bibitem[{Zoph et~al.(2016)Zoph, Yuret, May, and Knight.}]{multi_nmt_adapt}
Barret Zoph, Deniz Yuret, Jonathan May, and Kevin Knight. 2016.
\newblock Transfer learning for low resource neural machine translation.
\newblock \emph{EMNLP}.

\end{thebibliography}
\bibliographystyle{acl_natbib}

\newpage
\appendix
\section{\label{app} Appendix}

\subsection{\label{subsec:hyperparam}Model Details and Hyperparameters}
\begin{itemize}
    \item The LM similarity is calculated using a character-level LM\footnote{We sligtly modify the LM code from \url{https://github.com/zihangdai/mos} for our experiments.}
    \item We use character $n$-grams with $n=\{1,2,3,4\}$ for Vocab similarity and SDE. 
    \item During training, we fix the language order of multilingual parallel data for each LRL, and only randomly shuffle the parallel sentences for each language. Therefore, we control the effect of the order of training data for all experiments.
    \item For \disab-S, we search over $\tau=\{0.01, 0.02, 0.1\}$ and pick the best model based on its performance on the development set. 
\end{itemize}

\end{document}